\documentclass{article}

\usepackage{microtype}
\usepackage{graphicx}
\usepackage{subfigure}
\usepackage{booktabs} 
\usepackage[table,xcdraw]{xcolor}
\usepackage{bbm}
\usepackage{cuted}
\usepackage{flushend}
\usepackage{amsmath}

\usepackage{hyperref}
\hypersetup{
    colorlinks=true,
    linkcolor=blue,
    filecolor=green,      
    urlcolor=magenta,
}

\usepackage[accepted]{icml2019}

\icmltitlerunning{11-777 Project Proposal}

\begin{document}

\twocolumn[
\icmltitle{Multi-Modal Machine Learning for Assessing Gaming Skills in Online Streaming: A Case Study with CS:GO}



\icmlsetsymbol{equal}{*}

\begin{icmlauthorlist}
\icmlauthor{Longxiang Zhang}{}
\icmlauthor{Wenping Wang}{}
\end{icmlauthorlist}


\icmlkeywords{Multimodal, Twitch}

\vskip 0.3in
]




\begin{abstract}
Online streaming is an emerging market that address much attention. Assessing gaming skills from videos is an important task for streaming service providers to discover talented gamers. Service providers require the information to offer customized recommendation and service promotion to their customers. Meanwhile, this is also an important multi-modal machine learning tasks since online streaming combines vision, audio and text modalities. In this study we begin by identifying flaws in the dataset and proceed to clean it manually. Then we propose several variants of latest end-to-end models to learn joint representation of multiple modalities. Through our extensive experimentation, we demonstrate the efficacy of our proposals. Moreover, we identify that our proposed models is prone to identifying users instead of learning meaningful representations. We purpose future work to address the issue in the end.
\end{abstract}

\section{Introduction}
\label{intro}
Twitch.tv is an online video streaming platform that has witnessed fast growth in recent years\footnote{\url{https://www.newyorker.com/magazine/2017/11/20/how-to-get-rich-playing-video-games-online}}. It offers an open and live platform not only for people to have simple fun recording their gaming process, but also for potentially skilled gamers to showcase their gaming skills for possible talent scouting opportunities. This is especially true for popular e-sports that offer monetary tournaments and professional league careers. From an academic research point of view, Twitch.tv also presents a natural source of publicly available multimodal data (video, audio and chat) that is suitable for a gamut of deep learning tasks such as sentiment and emotion classification~\cite{leagueoflegends,ameer2022multi,zhang2022emotional}, video highlights~\cite{videohighlights_with_chat}, and spatial/temporal text grounding in video~\cite{textgrounding_in_video,yang2022tubedetr}. 

\begin{figure}[H]
    \centering
    \includegraphics[width=0.45\textwidth]{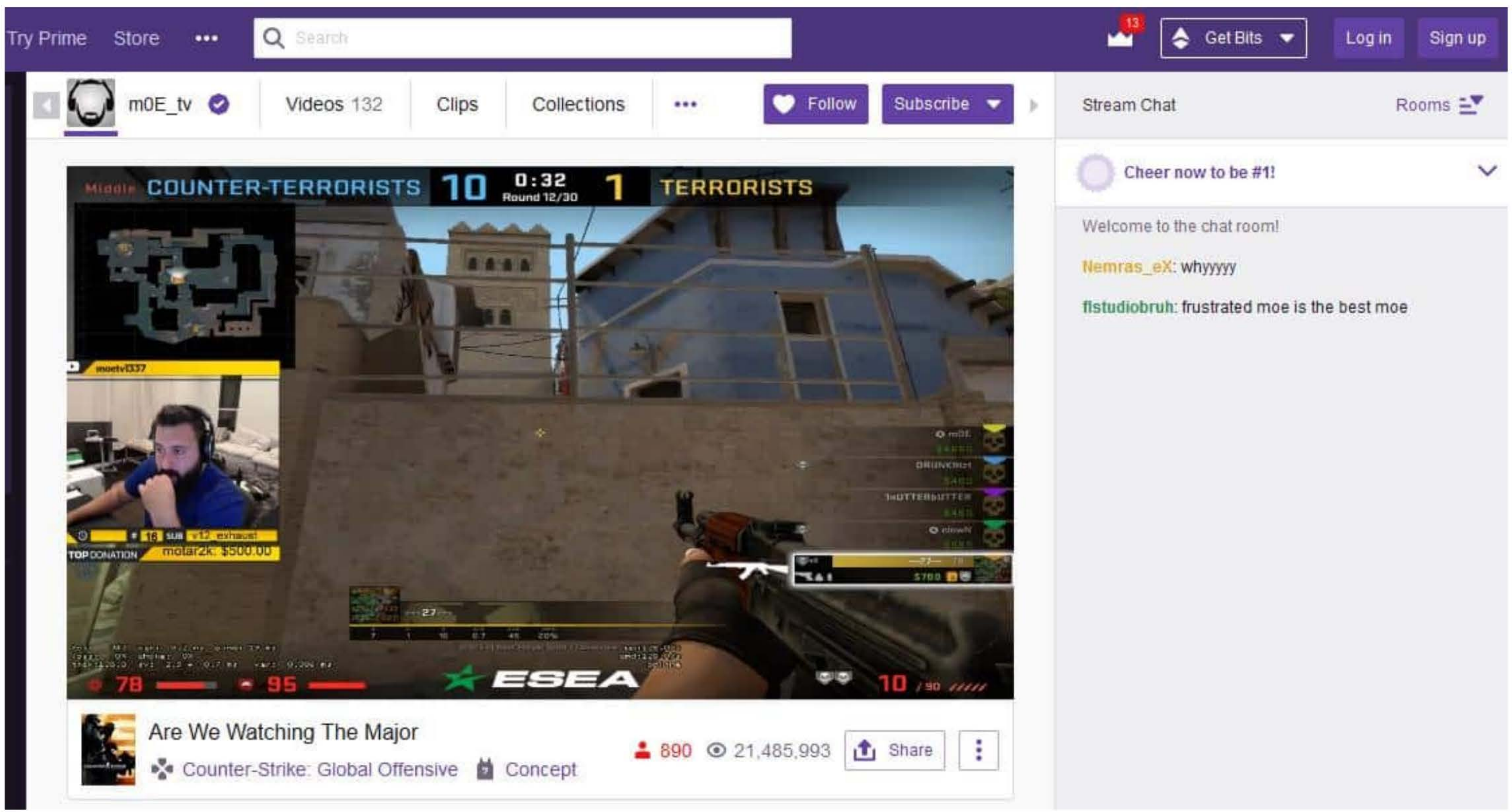}
    \caption{Twitch.tv graphical user interface}
    \label{fig:twitch_gui}
\end{figure}
Figure~\ref{fig:twitch_gui} shows a streaming for the game \textit{Counter Strike: Global Offensive} (CS:GO)\footnote{\url{https://blog.counter-strike.net/}}CS:GO. The left hand side of the screen shows the player's view and the player himself. The right hand side of the screen is a chat room where other twitch users can chat. Besides, viewers can hear sound effects of the game and the player's voice. Recently,~\citet{belova_e-sports_2019} created a multimodal dataset from Twitch.tv streams of the game CS:GO. The authors investigated evaluating gaming skills of players using Bayesian networks. However, they only carried out coarse grained analysis and their methodology was limited to shallow statistical models.

In this study, we seek to go beyond the approaches proposed in~\cite{belova_e-sports_2019}. Recent end-to-end models~\cite{lipreading}~\cite{tsai2019multimodal} have demonstrated success on multimodal classification tasks. Based on these models we investigated learning join/coordinated representation from multiple modalities that lead to better performance on evaluating gaming skills. 

Moreover, we carry out in-depth analysis of this dataset. Firstly, during data exploration we discover that many of the videos in the dataset are not for the CS:GO game. This issue is neglected by previous work. Secondly, we investigate utilizing numeric ranking in evaluating gaming skills. 

Last but not least, we empirically analyze how our deep models learn from the data. On the one hand,  there are only 1620 videos from 268 users. We empirically when making predictions our models tempt to utilize user identity instead of meaningful features. This is an interesting problem and worth investigating in future work. On the other hand, CS:GO videos are semi-structured. As shown in Figure~\ref{fig:twitch_gui}, the upper left part of gamers' view is a mini-map showing positions of him and his teammate, and the bottom left part shows gamers' health. We hypothesize that these parts contain useful information for evaluating gaming skills..

In summary, our contributions are:
\begin{enumerate}
    \item We manually clean the dataset by labeling each video with its corresponding game. Moreover, we demonstrate identity of users has strong impact on our model.
    \item We propose to learn coordinated representation of video modality and audio modality from CS:GO game videos. We also propose to utilize texts that are not temporally aligned with videos as prior information.  
    \item We investigate pre-training models with numerical rankings of users.
    \item We extract parts of screens with semantic meanings from videos and compare their ability in predicting gaming skills.
\end{enumerate}
The rest of this paper is organized as follows. Section 2 introduces related literature of our problem. Section 3 describes the proposed lipreading variations and the multi-view approach. Section 4 contains our experimental setup and section 5 contains the results. Section 6 concludes this paper and points out future direction.

\section{Related Work}
\label{related_work}
\subsection{Multimodal Approaches}


The work most closely related to ours is \cite{belova_e-sports_2019}, which introduced the dataset and a Bayesian network for evaluating gaming skills. They demonstrate that the latent values inferred with the Bayesian network are correlated with the actual ranking of players. However, they address the problem of evaluating gaming skills as a binary classification problem (A v.s. Non-A), whereas we consider all of the four ranks. Moreover, the representation they used for each modalities are extracted with pre-trained models and are simply concatenated. We hypothesize that joint/coordinated representations of multiple modality result in better performance.

Another related research area is audio-visual speech recognition (AVSR). Although the objectives are different, both our task and AVSR makes prediction based on the entire video clip. \citet{lipreading} propose a model for AVSR from raw video and audio, and we are using \cite{lipreading} as one of our baselines. A second similarly related work is \cite{soundnet} which again uses raw audio and video, but instead of doing classification builds good representations of sound in an unsupervised way.


\subsection{Video Modality}


There are a couple different areas of previous work focusing specifically on videos that are also related to our task. 

The first of these is video classification which is similar to our rank-section prediction task when using only the video modality. While most prior work focuses on explicitly classifying actions and we are not, we feel these methods will still be very relevant as the players skill is ultimately determined by the actions they take. Some notable works are as follows: \cite{two-stream_2014} introduced the two-stream architecture which is also used in \cite{belova_e-sports_2019}. \cite{karpathy_large-scale_2014} introduced the Sports-1M dataset in addition to several models capable of performing video classification over large datasets. \cite{ng_beyond_2015} presented methods for handling longer video clips. More recently, \cite{r2plus1d} and \cite{id3} have shown state of the art results on several datasets such as Sports-1M \cite{karpathy_large-scale_2014}, Kinetics \cite{kay_kinetics_2017}, and UCF101 \cite{soomro_ucf101:_2012}.


A second related area is that of video highlight prediction which is relevant in two ways: first, many prior works handling game streaming data were focused on the task of highlight prediction. Second and more importantly, we may use similar techniques as a type of hard attention to focus on the most important parts of the videos. Of particular relevance is \cite{unsupervised_highlight} since they do highlight prediction in an unsupervised way and we do not have labels for highlight prediction. Unfortunately \cite{unsupervised_highlight} also separate views of both the streamer and the game, so we may not be able to adapt their method directly. Some other relevant works in this category are \cite{gygli_creating_2014}, \cite{videohighlights_with_chat} -- which uses the chat data as supplementary information to determine the location of highlights -- and \cite{yahoo_esports_2016} -- which used the games special effects as a marker for the location of video highlights. Some inspiring work like ~\cite{yu2023inkgan} gives insights from another perspective for image treatment.

\subsection{Text Modality}

Online chat rooms for massive game streaming platforms have attracted academic interest in recent years \cite{leagueoflegends, twitchchat, arenasportsbar,prestridge2023play}. However, most literature work was focused on studying online chat language from a linguistic point of view \cite{arenasportsbar}, or simple statistical learning of the correlation between events in streaming video and chat topics \cite{leagueoflegends}. In \cite{twitchchat} it was shown that online chat language exhibits unique linguistic patterns or rules that are "short", "sudden", or even "convulsive", in addition to the irregular use of words and non-words (emojis, acronyms, internet slang and jargons) that don't conform to canonical English grammar. This presents a serious challenge for NLP neural network models~\cite{zhang2021leveraging, app13137753, yang2023linguistically} that are trained on clean English language. \\
In the deep learning community, there's a large body of research on studying a joint embedding using video and text modalities (\cite{textvideoembedding} and the references therein). Focal attention models \cite{focalattention_model,yang2022focal}, multi-view embedding \cite{multiview_embedding} and mixture of embedding expert model \cite{textvideoembedding} have all been proposed as novel network architectures to deal with challenges in joint representation learning (alignment, grounding, missing modality, etc.). However, most of the research is done on datasets with a strong semantic relation between the video and text modalities (e.g., text is describing the video). Online chat contents and game video streaming tend to be parallel and only weakly correlated with each other, as mentioned in \cite{leagueoflegends}, unless there are some "highlight events" (like one team close to winning the game), which usually incur a dominating topic in the chat room for a short period of time.

\section{Problem Statement}

The problem we addressed can be formalized as follows:
given a video with vision modality $X_V$, audio modality $X_A$, we want to predict the rank sections -- $\{A, B, C, D\} \in Y_R$. Additionally, for each user we might also have text infomation $X_C$. 


This is a standard multimodal classification problem where the classes are the rank sections (A, B, C, \& D) of the players

\section{Proposed Approach}

\label{model_description}

\subsection{Multimodal baseline models}
\label{baselines}
The baseline model we choose, based on our results from the midterm experiments, is adopted from the end-to-end Lipreading model for automatic speech recognition \cite{lipreading,han2023complex}.  This model achieved the highest accuracy among the other baseline models tried during our midterm. The model includes a spatio-temporal CNN, that convolves over time and spatial dimensions. A bidirectional GRU network is used for extracting features from video and audio modalities, and another additional bidirectional GRU for training a joint multimodal representation from both video and audio. A softmax layer is used on top of this joint representation for multi-class classification. The hyperparameters for the model architecture (E.g, kernel sizes, and padding in video CNN) are adopted directly from the paper. Cross entropy loss is used to train the model, where the classification problem is to predict/classify the ranking section of the player (4 classes), as shown in Eq. \ref{eqn:baseline3}. Adam optimizer was used for training with an initial learning rate of 0.0003.

\subsection{Lipreading Variations}

\begin{figure}
    \centering
    \includegraphics[width=0.45\textwidth]{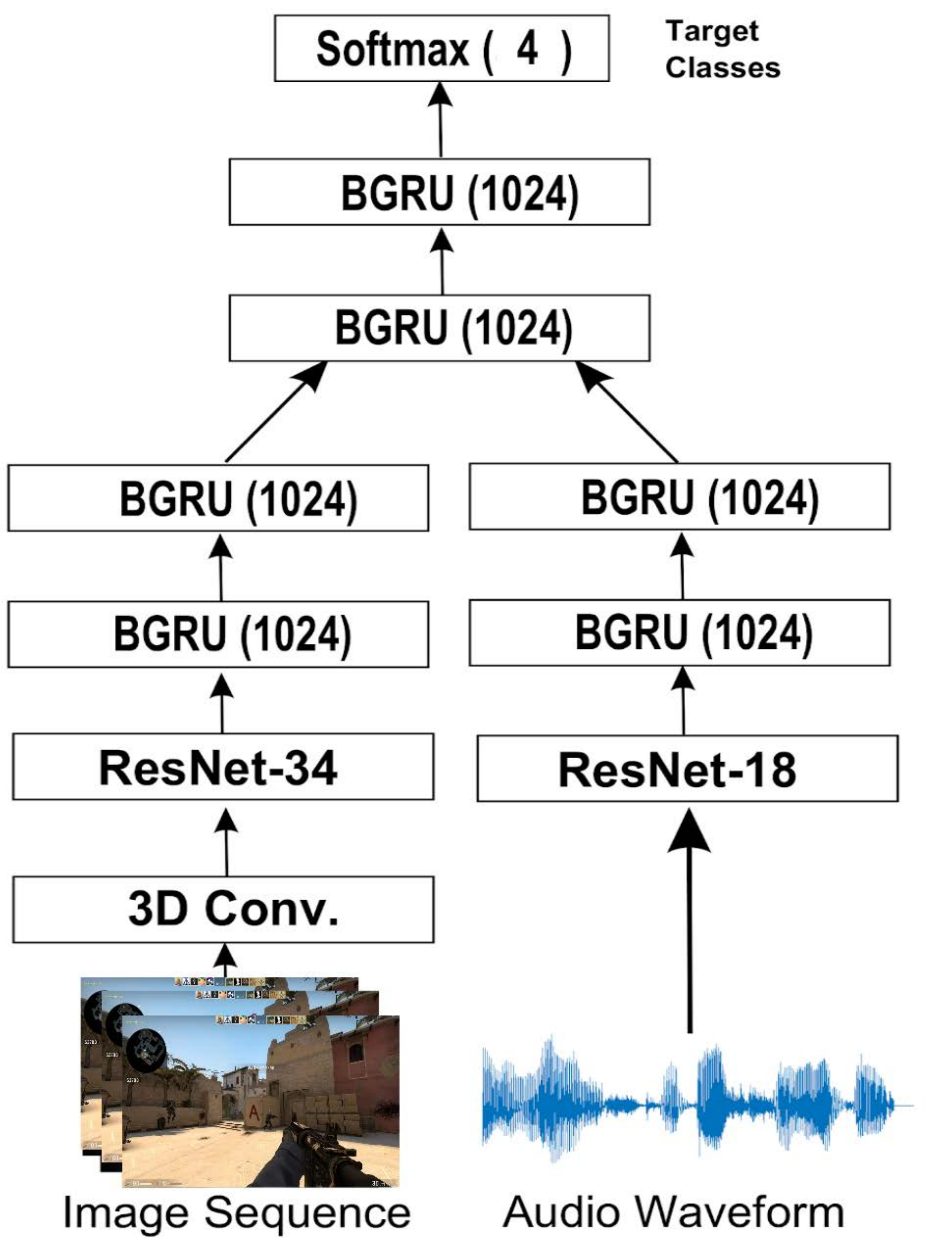}
    \caption{Diagram of the lipreading model}
    \label{fig:model_diagram}
\end{figure}

\begin{subequations}
\begin{align}
    \min_{\theta_V, \theta_A, \theta_{AV}} - \sum_{(x, y) \in D} \sum_{c \in \{0, 1, 2, 3\}} & \mathbbm{1}_{y = c} \log(\mathrm{Softmax}(f_{\theta_{AV}}(f_{\theta_V}(x_V), f_{\theta_A}(x_A)))) \\
    f_{\theta_V}(x_V) & = \mathrm{GRU}(\mathrm{Resnet}(\mathrm{CNN3D}(x_V))) \\
    f_{\theta_A}(x_A) & = \mathrm{GRU}(\mathrm{Resnet}(x_A)) \\
    f_{\theta_{AV}} (x) & = \mathrm{Softmax}( \mathrm{GRU}([x_A;x_V]))
\end{align}
\label{eqn:baseline3}
\end{subequations}

Many of the ideas we tried were extensions and modifications of this model. First we introduced a Kullback-Leibler (KL) divergence loss between the video and audio modalities on the outputs of their respective single modality parts of the model. In preliminary experiments the video modality gave significantly performance than audio and so our idea was that we could use the video to guide the training of the audio by forcing the audio to be in a similar space as the video. During training we only back-propagated the KL-loss through the audio part of the model as we did not want the audio to hurt the performance of the video.

\subsection{Multiview Model}

\begin{figure}
    \centering
    \includegraphics[width=0.45\textwidth]{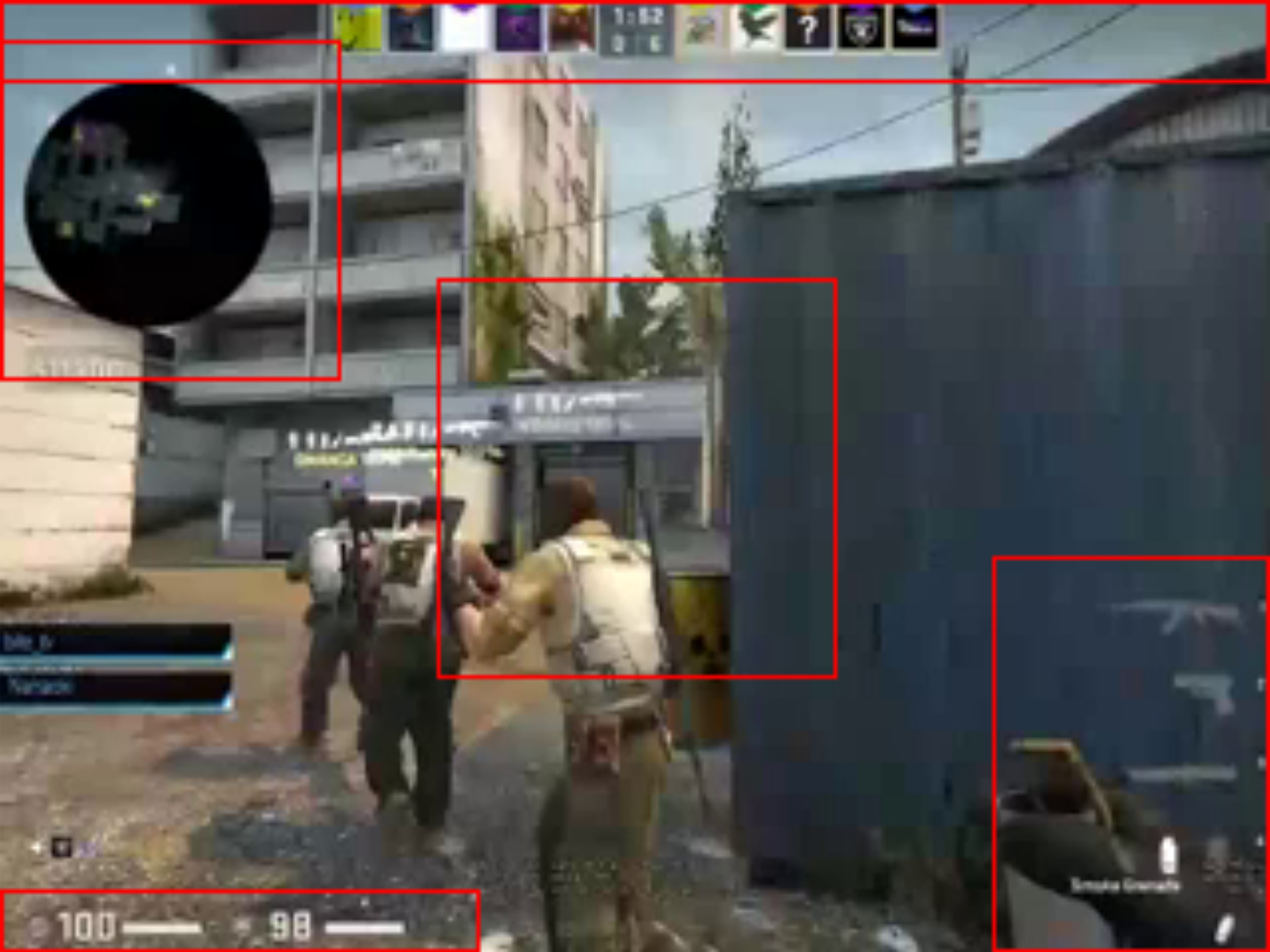}
    \caption{Red boxes show the different views we extracted from the frames}
    \label{fig:multiview_boxes}
\end{figure}

Another approach we tried was to use only the video modality, but to treat different parts of the frame as different views of the full video. The motivation behind this approach is that CS:GO contains UI elements that are in approximately the same location across all players and play sessions, and these UI elements contain semantic information that could be important for determining the players skill. For example how the players health changes over the course of the game. By processing the different UI elements or "views" separately we hoped to better learn the semantic information in each of the views separately before combining the information into a complete picture of the players skill. The different views we chose are shown in Figure \ref{fig:multiview_boxes} highlighted by red boxes. They are the minimap (top left corner), the top banner showing match scores and team information, the center where the player is aiming, the health bar (bottom left corner), and a view showing the player's selection of weapons with the currently equipped weapon highlighted (bottom right corner).

The model we use in this approach is very similar to the Lipreading model described previously. For the view specific models we use just the single stream part of the model before the concatenation step and add a final fully-connected layer to directly predict the player's rank section. For the full model containing all of the views, each view gets its own stream and then the output from all of the streams are concatenated together at the concatenation step.

\subsection{Text Prior}

For our final proposed idea we reintroduced the text modality into the model. The average number of messages for users ranked A, B, C and D are 1011, 1038, 350, and 851, respectively; and the total number of messages for those four ranks are 129104, 13126, 22677, 45278. Due to the fact that the chat data in the original dataset was not paired with the audio and video data we were not able use the chat data on a per video basis but instead aggregated all of the chats on each user's channel. We then used that chat data as a prior whenever predicting the rank for a video extracted from the associated user's channel. 
In order to create the chat prior we used a multilayered bi-directional LSTM to directly predict the user's rank section with all of the chats for that user's channel as input.\\ 
Similar to above mentioned split on video/audio data, we experimented with both message split and user split. However, due to the severe imbalanced data (some users have more than 10,000 chats while some have less than 10 corresponding messages), we decided to stick to message split for prior training. \\
We then tried two different ways of integrating the chat prior into the Lipreading model. First, we concatenated the chat prior with the final GRU output of the full model right before the final fully connected layer. Second, we concatenated the chat prior with the output from each of the single modality branches of the model before the final GRU.

\section{Experimental Setup}


\subsection{Dataset and Input Modalities}


\begin{table}[]
\centering
\caption{CS:GO vs Non-CS:GO}
\begin{tabular}{l|l|l}
          & Original & CS:GO Only \\
\hline
\# Videos & 3303     & 1620       \\
\# Users  & 748      & 268       
\end{tabular}
\label{tbl:dataset_stats}

\bigskip

\caption{Game type statistics}
\begin{tabular}{l|l}
Game Type            & \# Videos \\
\hline
CS:GO                & 1620      \\
Non-games            & 638       \\
First person shooter & 617       \\
Other games          & 428      
\end{tabular}
\label{tbl:game_stats}

\bigskip

\caption{Rank section statistics}
\begin{tabular}{l|l}
Rank Section & \# Videos \\
\hline
A            & 906       \\
B            & 315       \\
C            & 208       \\
D            & 191      
\end{tabular}
\label{tbl:rank_stats}
\end{table}

\paragraph{Dataset} The dataset~\cite{belova_e-sports_2019} was gathered from the Twitch live streaming platform.  It contains of videos of users playing CS:GO. Chat logs associated with games of the same users are also available, but they are on dates different from those of videos. The fact that texts are not temporarily aligned with videos prohibit us from utilizing three modality jointly. So in our models vision and audio modality are jointly used, and text modality are used if possible. In addition, metadata of streamers, including their follower count and their ranking in the E-sports Entertainment Association (ESEA) league are also available for some users. Specifically, users are classified into four groups (A, B, C and D) based on their gaming skills. Counts of users in each section are shown in table~\ref{tbl:rank_stats}. Besides, there are numerical rankings available. All videos come in length of 5 minutes with 10 FPS frame rate.  

\paragraph{Data Cleaning} We manually labeled the type of game being played in each video. As shown in table~\ref{tbl:dataset_stats}, we discovered that only half of the videos were about CS:GO and for more than two third of users there were no CS:GO videos. Table~\ref{tbl:game_stats} shows more details about type of games included in the dataset. So we decided to focus on a subset of 1620 videos and 268 users which were only for CS:GO. Statistics of ranks are summarized in table~\ref{tbl:rank_stats}.
\paragraph{Preprocessing} For vision modality, we reduced the video frame rate to 1 FPS so as to reduce memory footprint for training with the gain that the complete 5-minute video can be used in all our baseline models. In addition, videos were down-scaled from 1080p to 320*240 resolution. Audios are down-sampled with sampling rate 16000. We then extracted 20 Mel-frequency cepstral coefficients from them.  
\subsection{Data Split}
\label{sec:data_split}
After data cleaning, there are 1620 videos for the 268 users. Since our task is to predict the rank of of a gamer from his or her videos, with standard train-valid-test we randomly sample 80\% of videos as training set, 10\% of videos as validation set and 10\% of videos as test test. This data split is referred to as \textit{video-based split}. 

Meanwhile, videos from the same user may share some superficial characteristics that identify the user. Examples includes logos of users and their faces. User identity is a hidden co-variate for rankings and clearly a model should not predict user ranking based on it. To evaluate the extent to which our models utilize user identities, we carried out the same experiments on another data split we called \textit{user-based split}. We sample 80\% of users and use their videos as training set. The validation set and test set are constructed in the same way. As a result, videos in the three subsets are from different users.

In both video split and user split we ensure that the distribution of labels are the same in three subsets.

\subsection{Metric}
In our dataset, the distribution of class A, B, C and D is 56\%, 19\%, 13\%  and 12\%. So a naive majority voting can have 56\% of accuracy. To address this imbalance issue, we use precision, recall, and F-1 score to evaluate performance of our models.



\subsection{Hyper-parameters}
We adopt the model architecture with minimal changes to hyper-parameters. Convolution kernels are adjusted for our 320*240 input images, and input five-minute signals are temporarily down-samples to sequences of 30 hidden vectors. We use Adam as optimization algorithm and the learning rate is set to 0.0003. For the lipreading model, we adopt the original training strategy: first train models for each modality separately and then fine-tune the entire multi-modal model.

\section{Results and Discussion}
\label{results}
We have carried out extensive experiments on improving the Lipreading model and present all our results in Table \ref{tbl:results} (all rows starting with "LR"). All results presented in Table \ref{tbl:results} are weighted average across all four sections to account for the class imbalance issue. Majority class (first row) refers to a majority voting classifier as our baseline result. There are three major categories all our experiments fall into: (i) Variation on Lipreading (LR) model (ii) Introducing chat data as prior (iii) Breaking video frame into multiview streams. In the following subsections, we discuss detailed analysis on experiments within each category.
\begin{table}[]
\caption{Experimental Results}
\begin{tabular}{l|l|l|l}
\textbf{Model}        & \textbf{Precision} & \textbf{Recall} & \textbf{F1}    \\ \hline
Majority Class        & 0.410              & 0.640           & 0.500          \\ \hline
LR                    & 0.406              & 0.609           & 0.487          \\
LR+KL                 & 0.446              & 0.523           & 0.477          \\
LR+ranking loss       & 0.492     & 0.593           & 0.535 \\
LR+text before FC     & \textbf{0.535}              & 0.649           & 0.587          \\
LR+text before GRU    & 0.530              & \textbf{0.684}           & \textbf{0.597}          \\
\hline
LRA (video based)     & 0.557              & 0.599           & 0.510          \\
LRV (video based)     & 0.830              & 0.821           & 0.821          \\
LR (video based)      & 0.831              & 0.827           & 0.825          \\
LR+KL (video based)   & \textbf{0.852}     & \textbf{0.852}  & \textbf{0.850} \\
LR+mask (video based) & 0.493              & 0.593           & 0.531          \\  \hline
Center View           & 0.430              & 0.400           & 0.400          \\
Guns View             & 0.440              & \textbf{0.590}  & \textbf{0.500} \\
Health View           & 0.430              & 0.530           & 0.470          \\
Minimap View          & 0.400              & 0.250           & 0.290          \\
Top View              & \textbf{0.510}     & 0.480           & 0.490          \\
Multiview             & 0.470              & 0.470           & 0.470
\end{tabular}
\label{tbl:results}
\end{table}

\subsection{Variation on Lipreading Model}
\subsubsection{Kullback-Leibler Divergence Loss}
From the baseline results in \cite{belova_e-sports_2019} and our single modality results (LRA: Lipreading Audio branch, LRV: Lipreading Video branch, in Table \ref{tbl:results}), we've learned that video modality performs the best in single modality model across all metrics in player rank prediction. Therefore, simple concatenation of video and audio features is not guaranteed to lead to better results than single modality model. LR model already adopts two bidirectional GRU layers to integrate single modality features in a joint-representation like training framework, however, the effectiveness of such simple integration of unimodal features is questionable as can be seen from the marginal improvement in performance from LRV model to LR (video based) model in Table \ref{tbl:results}. We therefore introduced Kullback-Leibler divergence loss between the unimodal features from video and audio branch as additional objective during training. The KL divergence loss is calculated as $KL(f_{\theta_{V}}(x_{V}) || f_{\theta_{A}}(x_{A}))$, and backpropagated only through the audio branch in LP model. The motivation is to better utilize the best performing unimodal features (video) in guiding the feature extraction of audio features learned by audio branch. Introducing KL loss help improve model performance by a much larger margin ($\sim 2.5\%$) over unimodal models (LR+KL (video based) in Table \ref{tbl:results}).

\subsubsection{Video-based and User-based Data Split}
\begin{figure}
    \centering
    \includegraphics[width=0.45\textwidth]{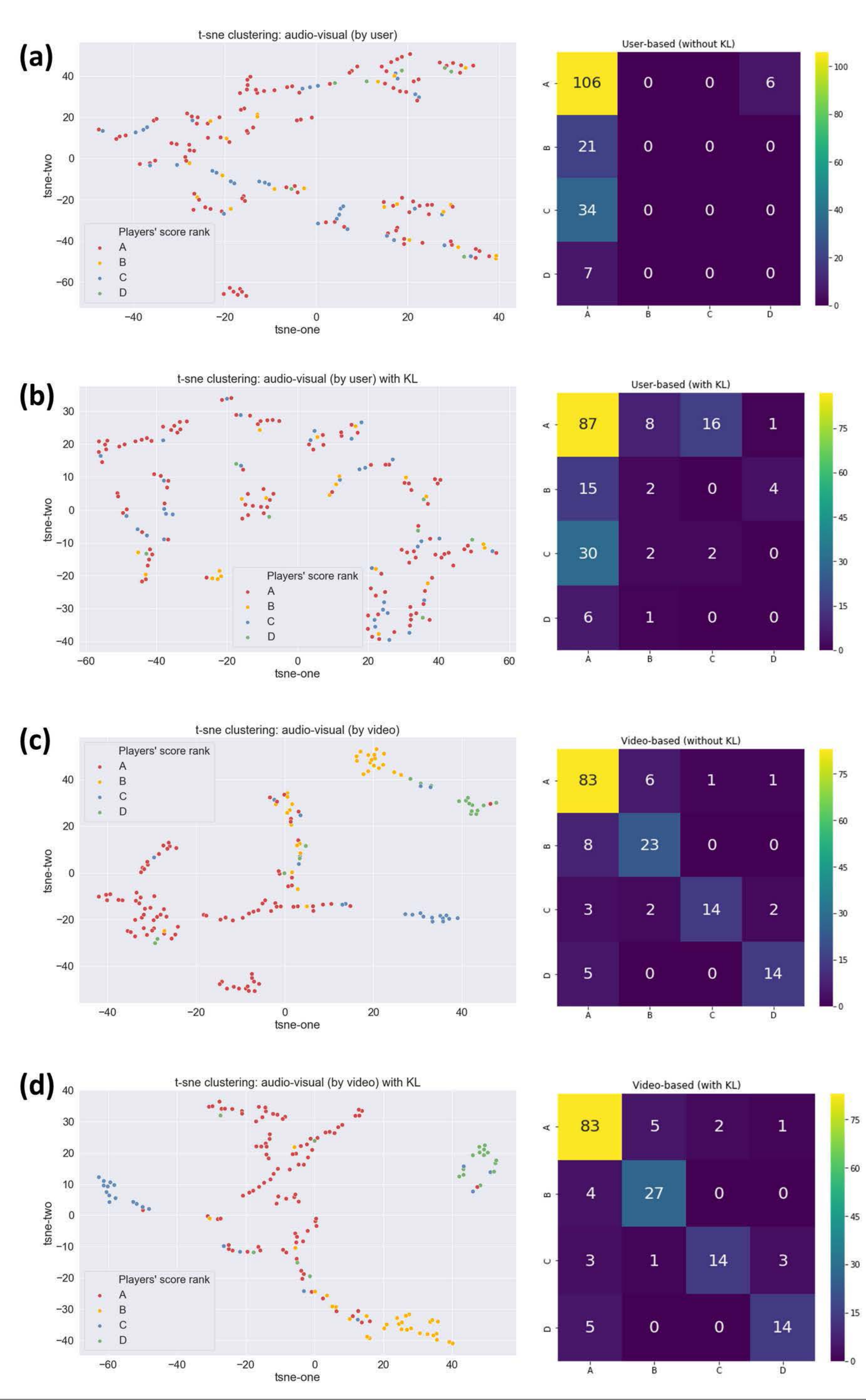}
    \caption{Visualization of multimodal features and confusion matrices. (a) LR (b) LR+KL (c) LR (video based) (d) LR+KL (video based). Four player sections are color coded in the TSNE plots (A: red, B: yellow, C: green, D: blue).}
    \label{fig:tsne_cm}
\end{figure}

Interestingly and confusingly, we observed a large discrepancy in model performance between two types of data split as described in Section \ref{sec:data_split}. All our models perform significantly better on \textit{video-based split} data than on \textit{user-based split}. As can be seen from Fig.\ref{fig:tsne_cm}, where we plot the confusion matrix from our prediction on test set and TSNE visualization of bimodal features extracted before the final FC layer in LP model. It's apparent that features learned on video-based split data exhibit much clearer clusters by player rank sections (different colors in Fig.\ref{fig:tsne_cm}), which is consistent with their much better classification performance. Since there is a significant overlap of users between training and test sets in video-based split dataset, we suspect that our model is picking up user identity information from one or both modalities. Audio stream data can contain user-specific data through typical voice of each user being recorded, but our single modality experiment on video-based data doesn't quite support this assumption as audio-only unimodal feature still performs the worst in classification task; interestingly, when we applied masks to video data to select only the center region from each video frame (about $80\%$ of the whole frame), the model performance on video-based split data drops significantly (LR+mask (video based) in Table \ref{tbl:results}) and becomes comparable to that of model trained on user-based split data. This shows that our LP model is picking up signals mostly from the edge region within each video for predicting player ranks, which motivates us to design the multiview model for in-depth study of dependence of model performance on different regions of videos, which is discussed in details in Section \ref{sec:multiview}.

\subsubsection{Ranking Loss Pretraining}
Training on user-based split data not only suffers from weak performance, but also unstable training. As can be seen from Fig.\ref{fig:tsne_cm} (a), training multimodal LP model from scratch can sometimes cause the model to collapse into a simple majority voting classifier. We therefore experimented with a lot of different techniques in order to stabilize model training and improve model performance: upsampling the minority class in training, introducing weights to cross-entropy loss, adding KL divergence loss with hyperparameter-tuned learning rate (0.001) and scale (1), and pretraining unimodal branches with different objective such as ranking loss. The pairwise ranking loss is defined as follows:

\begin{equation}
\label{eqn:ranking_loss}
    L(f'_V(x_{1V}), f'_V(x_{2V})) = \max (0, -y(f'_V(x_{1V})-f'_V(x_{2V}))+m)
\end{equation}

$f'$ here refers to the video (audio) branch with additional fully connected and softmax layers for calculating player rank logits. $y$ is $+1$ for pairs of videos (audios) in the same section and $-1$ for pairs from different sections, $m$ is the margin (0.2 in our experiments). Video (audio) pair data is generated based on user-based split, and we do subsampling of the pair data at $10\%$ to maintain the size of training set due to time and resource limit.

The ranking loss pretraining for video (audio) branch not only stabilizes training of the full model, but also improves the model performance in precision and F1-score (LR+ranking loss in Table \ref{tbl:results}) by large margin. We choose not to present all results from other aforementioned techniques due to space limit and comparable results. These experiments indicate that a better model performance may be achieved with pretraining of suitable objective.

\subsection{Text prior}
\begin{figure}
    \centering
    \includegraphics[width=0.45\textwidth]{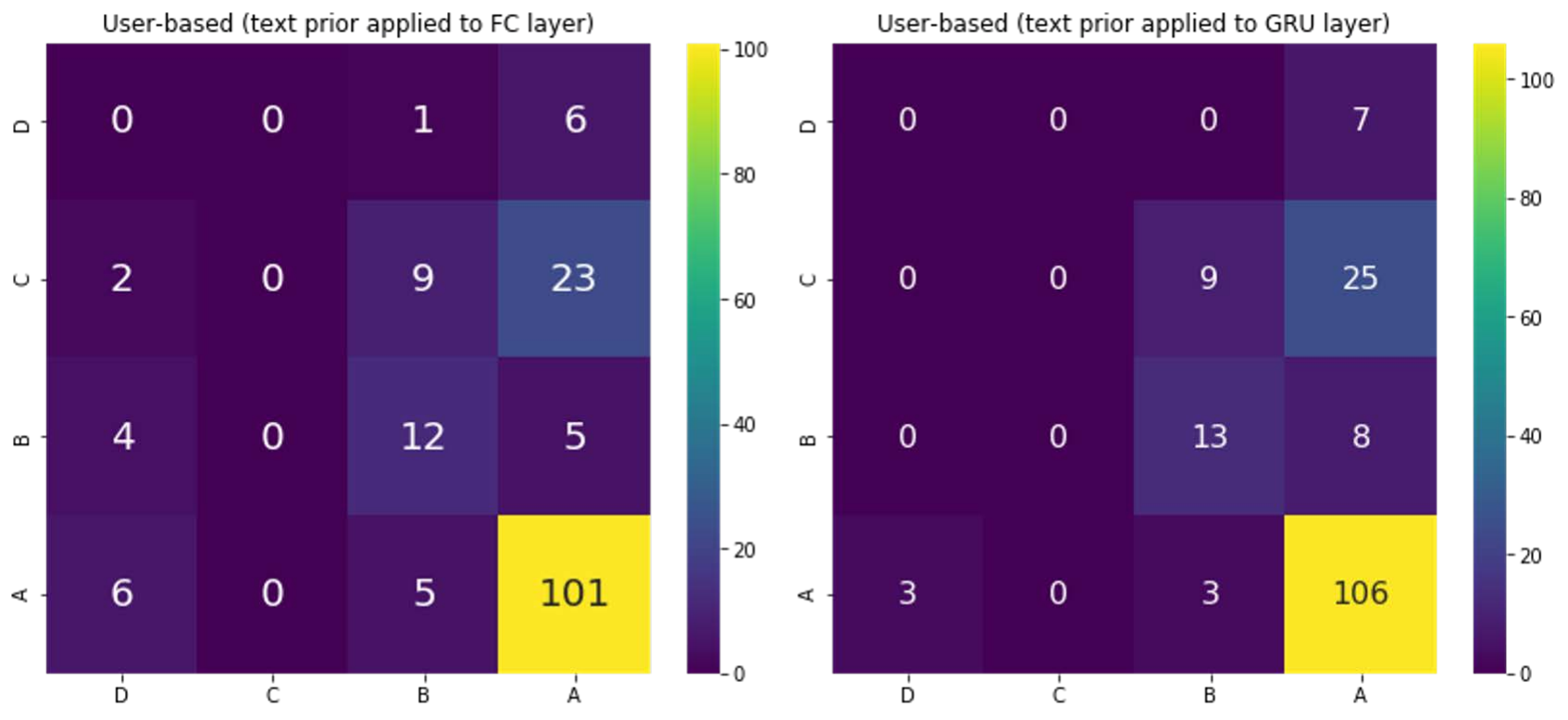}
    \caption{Confusion matrices for adding text features as prior. \textbf{Left:} text features added before FC layer; \textbf{Right:} text features added before GRU layers. Rank sections are (top-down and left-to-right) D, C, B and A.}
    \label{fig:text_prior}
\end{figure}

"LR+text before FC" and "LR+text before GRU" in Table \ref{tbl:results} show the model performance on user-based split data when incorporating text prior for each user. Since the text features are pretrained on the same prediction task, we initially conjectured that adding text prior to the FC layer should be most effective; interestingly, the results indicate that adding the text prior before the final integration GRU layers actually leads to better performance, although our incorporation of text features into GRU layer is simple (repetition and concatenation). However, the numerical results for metrics may be misleading as can be seen from Fig.\ref{fig:text_prior}: although the F1 and Precision scores are improved by adding text prior, the model fails to predict correctly any video from rank section C and D. We are not sure why inclusion of the text prior could push the model towards better differentiation only between the majority classes (A and B). This is probably not caused by class imbalance issue as the number of chat sentences either per section or per user are quite balanced in our dataset. The results nonetheless are encouraging in that it proves effectiveness of chat modality even without alignment in player rank prediction, when used properly in the model.

\subsection{Multiview model}
\label{sec:multiview}

In Table \ref{tbl:results} we also present results for our multiview experiments, both when using each of the views individually, and when using all of the views together. We find that although the Top View provides the highest precision, the Guns View gives better recall and a better overall F1 score. It is unclear exactly why the full multiview setting would perform worse than the top view and guns view individually, but our hypothesis is that some of the other views -- such as the minimap view -- may be introducing noise and thus hindering learning. While the best single view model did outperform the lipreading baseline, it did not outperform the majority class baseline, and the full multiview model did worse than both. One reason for this poor performance could be that the specific views we chose are missing important and relevant information learning the player's rank. A second reason could be because of how the semantic information from each pipeline is siloed until a later part of the model. It could be that in order for the model to interpret one view correctly it needs information from some other view or views earlier on. For example interpreting the low level pixel information of the minimap may be helped by seeing the center view containing information about what the player sees in that portion of the map. Finally, although in previous experiments the audio-only modality gave poor results, its possible that removing the audio completely hurts performance as well.

\section{Conclusion and Future Directions}

We have showed significant progress towards our goal of predicting player skill directly from Twitch streaming data through improving the dataset, introducing KL and ranking losses, integrating a chat prior, and learning on multiple semantically relevant views. However, we are still far from our goal of producing a high performing model that could reasonably be used in the wild.

Our data cleaning process has not only improved the quality of the datasets, but also introduced additional training objectives (e.g. type of games based on video/audio) that can be explored on current dataset. Another interesting direction for experiment is to reframe player rank prediction not as a classification problem, but a ranking prediction problem where the objective is to predict whether one video is showing better skills than another. Regression-like model widely used in other domains~\cite{ye2017sparse, yue2018deep} can also be derived if we focus only on one rank section and design model for prediction player position.

The multiview experiment, together with ablation study done on unimodal branches of our model, indicates that our model performance may be determined by certain views within each video stream instead of the whole stream. Studying how to optimally select such informative views within video and audio streams through supervised or unsupervised learning may be interesting and challenging topic for future research work. Adapting recent advances of accelerator hardware to the models~\cite{wang2023integrity, yuan2018rosetta, 10.1145/3174243.3174255} could be another direction to explore.

Additionally, we feel that one of the major limitations of this work was the size and quality of the dataset and a major improvement over this work would be to expand the current dataset with more relevant and clean video/audio data and inclusion of chat data that is properly aligned with the video data. An improved dataset would greatly reduced the problems we had dealing with both class imbalances and overfitting.

\nocite{langley00}

\bibliography{references}
\bibliographystyle{icml2019}

%
%



\end{document}